\documentclass{article}
\usepackage{spconf,amsmath,graphicx}
\usepackage{indentfirst}
\usepackage[pagebackref=true,breaklinks=true,letterpaper=true,colorlinks,bookmarks=false]{hyperref}
\usepackage{mathtools}

\usepackage{amssymb}
\usepackage{multirow}
\usepackage{siunitx}

\usepackage{color, colortbl}
\definecolor{LightCyan}{rgb}{0.88,1,1}
\definecolor{LightRed}{rgb}{1,0.88,0.95}
\usepackage[boxed]{algorithm2e}
\SetAlCapSkip{1em}
\usepackage{comment}
\usepackage{amsmath}
\DeclareMathOperator*{\argmin}{\arg\!\min}

\SetKw{Continue}{continue}
\title{False Positive Removal for 3D vehicle detection \\with penetrated point classifier}
\name{Sungmin Woo\sthanks{E-mail: smw3250@yonsei.ac.kr}, Sangwon Hwang, Woojin Kim, Junhyeop Lee, Dogyoon Lee, and Sangyoun Lee\sthanks{Corresponding Author, E-mail: syleee@yonsei.ac.kr}}
\address{School of Electrical and Electronic Engineering, Yonsei University, Seoul, Korea}
\begin{document}
	\ninept
	\maketitle
	\begin{abstract}
		
		Recently, researchers have been leveraging LiDAR point cloud for higher accuracy in 3D vehicle detection. Most state-of-the-art methods are deep learning based, but are easily affected by the number of points generated on the object. This vulnerability leads to numerous false positive boxes at high recall positions, where objects are occasionally predicted with few points. To address the issue, we introduce Penetrated Point Classifier (PPC) based on the underlying property of LiDAR that points cannot be generated behind vehicles. It determines whether a point exists behind the vehicle of the predicted box, and if does, the box is distinguished as false positive. Our straightforward yet unprecedented approach is evaluated on KITTI dataset and achieved performance improvement of PointRCNN, one of the state-of-the-art methods. The experiment results show that precision at the highest recall position is dramatically increased by 15.46 percentage points and 14.63 percentage points on the moderate and hard difficulty of car class, respectively.
	\end{abstract}

	\section{Introduction}
	\label{sec:intro}
	3D object detection is a crucial task in autonomous driving since vehicles must be provided with accurate scene information. It is essential for safe driving to recognize the other cars driving ahead or parked at the side of the road precisely. To achieve this goal, recent works of 3D object detection utilize the data from several types of sensors, such as a monocular camera, stereo cameras, and LiDAR. They provide various and useful data, but each sensor has its limitations.
	
	RGB images from cameras contain appearance information that can be efficiently exploited by the advanced techniques of 2D Convolutional Neural Network (CNN). However, they lack depth information which is material to 3D object detection. This drawback results in considerably low performance for monocular 3D object detection \cite{naiden2019shift, Wang_2019_CVPR, Chen_2016_CVPR, Mousavian_2017_CVPR}, that only uses a single image to detect the objects in 3D space. To overcome the limitation, \cite{you2019pseudo, li2019stereo, chen2020dsgn} used two images from the stereo camera to estimate the depth of the scene more precisely, by sharing the weights of image feature extractors. The estimated depth of the stereo images is more enhanced than monocular images, but due to the discrete nature of the pixels, they still are insufficient for real-world applications. Therefore, depth prediction from 2D images itself is not adequate for autonomous driving, where an accurate location of the object is required.
	
	On the other hand, point clouds from LiDAR provide precise distance information, which makes it easy to capture the spatial structures of the scenes. Thus, most 3D object detection methods with high performance are point-based detection. However, the LiDAR point cloud has fatal weaknesses of being sparse in the long range, and that the points are irregular and unordered. To handle these challenges, existing methods such as PointNet~\cite{Qi_2017_CVPR} and PointNet++~\cite{qi2017pointnet++} directly process the raw point cloud regardless of the order of the points. 
	Alternatively, point clouds are converted into specific forms that can be easily processed by projecting a point cloud onto the corresponding image for 2D CNN~\cite{ku2018joint, yang2018pixor, liang2018deep, chen2017multi} or dividing into voxels for 3D CNN~\cite{zhou2018voxelnet, song2016deep, yan2018second}.
	
	\begin{comment}
	\begin{figure}[t]
	\begin{center}
	\includegraphics[width=1\linewidth]{images/DD.pdf}
	\end{center}
	\caption{Overview of our Shift R-CNN hybrid model. Stage 1: Faster R-CNN with added 3D angle and dimension regression. Stage 2: Closed-form solution to 3D translation using camera projection geometric constraints.
	}
	\label{fig:our_3stage_system}
	\end{figure}
	\end{comment}
	\begin{figure*}[t]
		\begin{center}
			\includegraphics[width=1.0\linewidth]{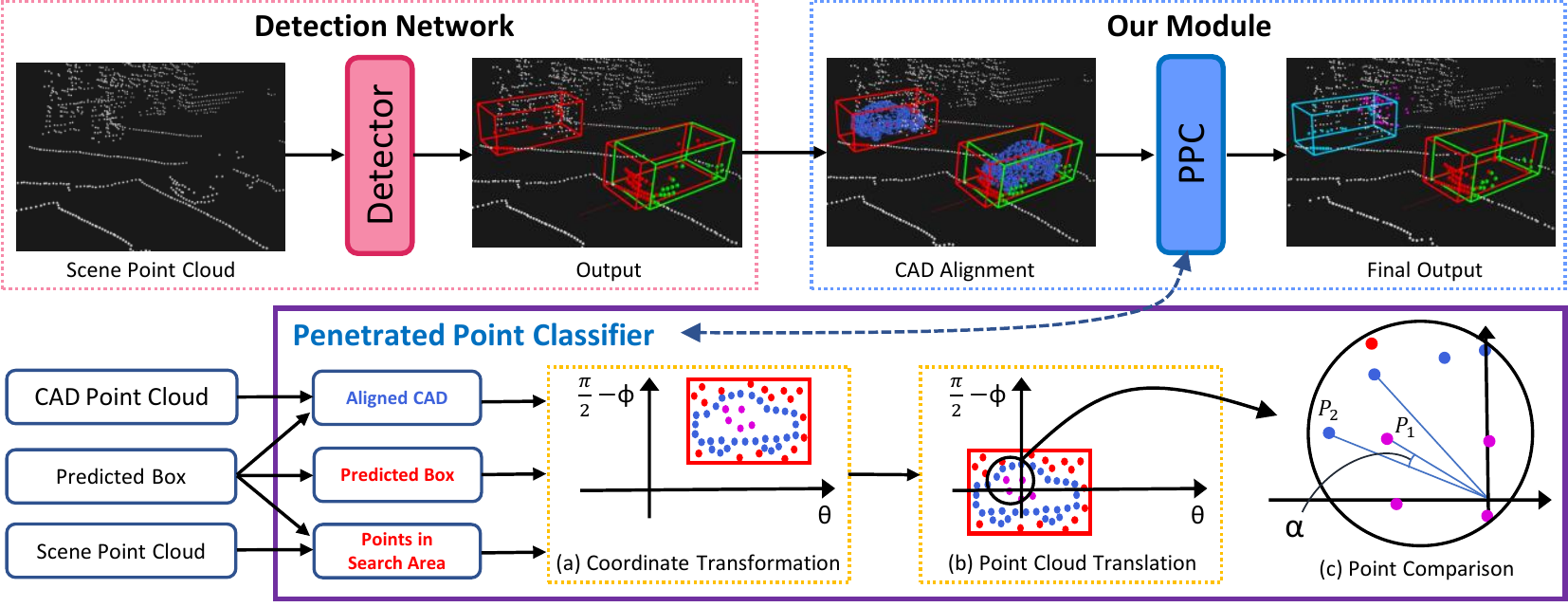}
		\end{center}
		\vspace{-0.6cm}
		\caption{Overview of our PPC module in the overall detection pipeline. The framework is divided into three stages. Aligned CAD point clouds, predicted boxes, and search area points are obtained in advance as described in Sec.~\ref{sec:method}. (a) Representation of 3D spherical coordinate on 2D Cartesian coordinate without radial distance. (b) Translation of inputs to origin. (c) Comparison of 2D polar coordinates of $point_{SA}, P_1,$ and $point_{CAD}, P_2$. The point in search area, $point_{SA}$, is determined to be the penetrated point according to Algorithm.~\ref{alg:ppc}. The predicted box is in red, ground-truth box is in green, CAD point cloud is in dark blue, false positive box is in light blue, and penetrated points are in pink.
		}
		\vspace{-0.4cm}
		\label{fig:ppc}
	\end{figure*}

	Based on these approaches, significant progress has been made in how the point cloud can be effectively utilized from the limited information it provides. Nevertheless, due to the sparsity of point clouds and frequent occurrence of occlusion, it is still challenging to determine whether a set of points represents an actual object or not, when too few points are generated on the object. Moreover, advanced methods using deep learning networks also suffer from solving this problem with scanty information provided. As a result, many false positive boxes are generated with the existing methods, causing a decrease in performance. This is also a critical issue in autonomous driving. The autonomous vehicles must recognize as many objects in the scene, which means that the model with high recall position should be applied. However, depending on the score threshold of the final prediction from the network, precision gets lower as recall gets higher. As the high recall is necessary but the precision gets lower, increasing the precision at high recall position is one of the main issues~\cite{seif2016autonomous, wu2017squeezedet, xiang2019comparative} that need to be solved in autonomous driving. We are demanding that besides talking about the average precision of the detectors, the precision at high recall position should be addressed as a separate problem for the actual application.
	
	In this paper, we propose a new approach in 3D vehicle detection for the false positive removal task. False positive boxes tend to be generated in the area with sparse points, but the neural network needs a certain number of points that contain information where features can be extracted. Thus, we focus on configuring a simple but intuitive module which can effectively sort out the false positive boxes, no matter how sparse points are in the predicted box. The proposed method increases the precision substantially at the high recall position, which indicates that numerous false positive boxes are removed from the final predicted boxes while preserving true positive boxes.

	\section{METHOD}
	\label{sec:method}
	\vspace{-0.3cm}
	\subsection{Overview}
	\label{overview}
	
	The basis of the proposed approach is the fundamental property of LiDAR that the laser cannot penetrate the objects. It can be inferred from the property that if any penetrated point is generated in the space behind the vehicle of the predicted box, the box must be recognized as misprediction. In other words, we assume that the penetration can only occur for false positive boxes. To verify the assumption, the boundary of the space behind the vehicle, which will be termed \textit{penetrated area}, should be fixed. However, there is an inevitable problem that it is impossible to figure out the exact shape of a vehicle inside the predicted box with few points on it. We resolve the problem by using a point cloud of Computer-Aided Design (CAD). In our module, a certain CAD point cloud of a car with generalized shape is used to replace the unattainable point cloud of an actual vehicle, in which points are only located at 3D contour. Generalizing different shapes of vehicles into the identical CAD model can be an issue, but we tackle it in Table.~\ref{tab:model} by experiments.
	
	After the shape of a vehicle in the predicted box is gained by the CAD model, the remaining task is to determine whether the points exist in the penetrated area. We introduce the Penetrated Point Classifier (PPC) in Sec.~\ref{sec:ppc}, which can efficiently deal with the point cloud in the 3D space. By using the PPC, the penetrated area is set from the view of LiDAR, and the point cloud is transformed from the 3D cartesian coordinate into the 2D polar coordinate to determine whether the penetration occurs undoubtedly.

	% 상원이가 써본거
	\vspace{-0.2cm}
	\subsection{CAD Point Cloud Alignment}
	\vspace{-1.5mm}
	\label{sec:CAD}
	First, we pre-process the raw CAD point cloud of a car with a generalized model. Since the model is only required to have an adequate number of points to represent the shape of a car, it is down-sampled to $N_s = 500$ points for low complexity. Let $\textbf{P} = {\{\textbf{p}_{(i)}=[x_{(i)},y_{(i)},z_{(i)}]}\}_{i=1,\dots,N_s} \in \mathbb{R}^{N_s\times3}$ be the CAD point cloud and $\textbf{S}=[w, l, h] \in \mathbb{R}^{1\times3}$ be the size of the predicted box or CAD model. The aligned CAD point cloud $\textbf{P}_{ac}$ could be formulated as follows:
	\begin{gather}
	\label{eq1}
	\textbf{P}_{sc} = \kappa (\textbf{S}_{b} \oslash \textbf{S}_{c}) \odot \textbf{P}_{c},
	\end{gather}
	\begin{gather}
	\label{eq2}
	\textbf{P}_{rsc} = \textbf{P}_{sc} 
	\begin{bmatrix} \cos(\theta) & -\sin(\theta) & 0 \\ \sin(\theta) & \cos(\theta) & 0 \\ 0 & 0 & 1 \end{bmatrix}^T,
	\end{gather}
	\begin{gather}
	\label{eq3}
	\textbf{P}_{ac} = \textbf{P}_{rsc} + \textbf{u}_{bc}.
	\end{gather}
	where $\oslash$ and $\odot$ mean element-wise division and multiplication, $\theta$ is the box orientation from the bird's view, $\textbf{S}_{b}$ and $\textbf{S}_{c}$ are sizes of the predicted box and CAD model, and $\textbf{u}_{bc} \in \mathbb{R}^{1\times3}$ is the center coordinates of the predicted box. $\textbf{P}_{c}$, $\textbf{P}_{sc}$ and $\textbf{P}_{rsc}$ are the point clouds of CAD model centered at the origin, scaled with the predicted box, and rotated and scaled CAD model, respectively. The ratio of CAD model $\kappa$ is the hyperparameter that adjusts the size of the CAD model over the size of the predicted box. The importance of the hyperparameter $\kappa$ is described in Sec.~\ref{sec:eval}.
	
	The CAD point cloud is scaled and rotated relative to the predicted box in Eq.~\ref{eq1} and Eq.~\ref{eq2}, and translated to the position of the predicted box in Eq.~\ref{eq3}. Each predicted box will have its aligned CAD point cloud inside, which is then used as the input to the Penetrated Point Classifier.

	\begin{figure}[t]
		\begin{center}
			\includegraphics[width=1\linewidth]{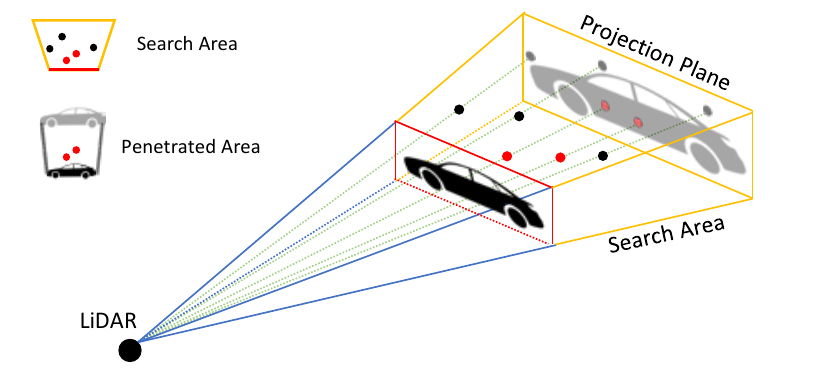}
			
		\end{center}
		\vspace{-0.5cm}
		\caption{Illustration of the search area and the penetrated area. The virtual projection plane is drawn to facilitate understanding. All points are in the search area while only red points are in the penetrated area.
		}
		
		\label{fig:area}
		\vspace{-0.3cm}
	\end{figure}
	
	\begin{table*}[t]
		\small 
		\vspace{-2mm}
		\begin{center}
			\scalebox{1}[1]{
				\setlength\tabcolsep{10pt}
				\begin{tabular}{|c||cc|cc|cc|cc|}
					\hline
					\multirow{3}{*}{Method} & 	
					\multicolumn{4}{c|}{~~Car - 3D Detection ~~} & \multicolumn{4}{c|}{~Car - BEV Detection~} \\
					\cline{2-9}
					& \multicolumn{2}{c|}{HR-Precision} & \multicolumn{2}{c|}{Average Precision} & \multicolumn{2}{c|}{HR-Precision} & \multicolumn{2}{c|}{Average Precision} \\
					%\cline{2-9}
					& Mod. & Hard & Mod. & Hard & Mod. & Hard & Mod. & Hard\\
					\hline
					
					PointRCNN & 46.59 & 53.90 & 82.26 & 77.99 & 51.79 & 59.00 & 88.89 & 86.48 \\
					PointRCNN + PPC (Ours) & \textbf{62.05} & \textbf{68.53} & \textbf{82.65} & \textbf{78.36} & \textbf{69.44} & \textbf{75.27} & \textbf{89.35} & \textbf{86.76} \\
					%\rowcolor{LightCyan}
					\textit{Improvement} & \textit{+15.46} & \textit{+14.63} & \textit{+0.39} & \textit{+0.37} & \textit{+17.65} & \textit{+16.27} & \textit{+0.46} & \textit{+0.38} \\
					\hline
				\end{tabular}
			}
		\end{center}
		\vspace{-0.4cm}
		\caption{Performance comparison on KITTI \textit{val} set. The HR-Precision and AP are evaluated with 40 recall positions.
		}
		\label{tab:val}
		\vspace{-0.2cm}
	\end{table*} 
	\vspace{-0.2cm}
	\subsection{Cropping Search Area}
	\label{sec:crop}
	The PPC performs binary classification on points of the scene, regarding whether they are positioned in the penetrated area. Since the task is to verify the presence of a penetrated point, it is unnecessary to classify all points in the scene. To ease the computation, we crop the point cloud behind the predicted box, which will be termed \textit{search area}. It is the most appropriate area to be examined because the search area contains the penetrated area with a similar scale, as the predicted box contains the predicted vehicle. The search area and the penetrated area are depicted in Fig.~\ref{fig:area}, and the method to crop the points in the search area is described in Sec.~\ref{sec:ppc}.

	\subsection{Penetrated Point Classifier (PPC)}
	\label{sec:ppc}
	
	The inputs of the Penetrated Point Classifier consist of the predicted boxes, aligned CAD point clouds, and points in the search area, as mentioned in Sec.~\ref{sec:CAD} and Sec.~\ref{sec:crop}. To gain the points in the search area, the point cloud must be considered from the view of the LiDAR. As seen in Fig.~\ref{fig:area}, the search area has the shape of a frustum in the 3D cartesian coordinate, where the width gets bigger as it gets more distant from the LiDAR. Since this representation makes it hard to classify the points, we propose to transform point cloud data from the 3D cartesian coordinate which is $(x, y, z)$ to spherical coordinate which is $(r, \theta, \phi)$. $r$ is the radial distance, $\theta$ is the azimuthal angle, and $\phi$ is the polar angle in 3D space. Regardless of the value of the radial distance $r$, the point cloud is then projected to the 2D plane of a 2D cartesian coordinate where a horizontal axis is $\theta$, the vertical axis is $\phi$, and an origin point is LiDAR, like a range image which can be seen in Fig.~\ref{fig:ppc}-(a). As we can infer from the figure, points in the search area should satisfy the following inequalities:
	\begin{gather}
	\label{eq4}
	\max(r_{box}) < r_{SA},\nonumber\\ \min(\theta_{box}) < \theta_{SA} < \max(\theta_{box}),\\ \min(\phi_{box}) < \phi_{SA} < \max(\phi_{box})\nonumber
	\end{gather}
	where $(r_{SA}, \theta_{SA}, \phi_{SA})$ and $(r_{box}, \theta_{box}, \phi_{box})$ are the 3D spherical coordinates of the points in search area and on the predicted box. The area covers slightly more than the actual space behind the vehicle when the box is rotated, but this unexacting representation is reasonable in terms of complexity since it simply is the search area.
	
	To classify whether a point in the search area is in the penetrated area, the boundary of the CAD point cloud should be distinct. However, the CAD point cloud is comprised of the points, not a line, making the boundary obscure. We deal with this problem by translation of inputs and transformation of the coordinate system once more, as shown in Fig.~\ref{fig:ppc}-(b), (c). The point cloud on the 2D plane is translated to the origin, and the coordinate system is transformed from 2D cartesian coordinate which is $(\theta, \phi)$, to 2D polar coordinate which is $(\rho, t)$. $\rho$ is radial coordinate and $t$ is angular coordinate in 2D plane. In the 2D polar coordinate system, the PPC searches a point of CAD point cloud which has the closest angular coordinate to that of an interested point in the search area. For example, in the Fig.~\ref{fig:ppc}-(c), $\alpha$ is the minimum gap between the angular coordinates of the interested point $P_1$ in the search area and the points of the CAD point cloud. The point $P_2$ then becomes the point of the CAD point cloud that has been searched. The PPC compares the radial coordinates of the points, $\rho_{P_1}$ and $\rho_{P_2}$, and if $\rho_{P_1}$ is smaller than $\rho_{P_2}$, the interested point $p_1$ is considered as the penetrated point. The rationale behind this logic is that the CAD point cloud of a car is composed of the points that are closely located and has a shape of the closed surface with a gentle gradient in the view of the center. The formulation of the PPC is summarized in Algorithm.~\ref{alg:ppc}.
	\setlength{\textfloatsep}{8pt}
	\begin{algorithm}[t]
		
		\KwIn{points in aligned CAD point cloud $(AC)$, \hspace{1.5cm} points in search area $(SA)$, \hspace{2cm} predicted boxes $(box_{pred})$}
		\KwOut{final predicted boxes $(box_{final})$}
		{\bf Initialization:} $box_{final} \gets box_{pred}$ 
		
		\vspace{0.2cm}Do coordinate transformation and translation\\according to Sec.~\ref{sec:ppc}.
		\vspace{0.3cm}
		
		\For{$box_i \in box_{pred}$}{
			\vspace{0.1cm}
			\For{$point_{SA} \in SA$}{
				\vspace{0.1cm}
				\hspace{-0.1cm}$point_{CAD} \gets \argmin\limits_{point_{CAD}} | t_{point_{SA}} - t_{point_{CAD} \in AC} |$
				\vspace{0.05cm}
				
				\If{$\rho_{point_{CAD}} > \rho_{point_{SA}}$}{
					\vspace{0.2cm}Delete $box_i$ from $box_{final}$
					
					\textbf{break}
					\vspace{0.1cm}
			}}
		}
		\textbf{return} $box_{final}$
		
		\label{alg:ppc}
		\caption{PPC Formulation}
	\end{algorithm}
	\begin{comment}
	\begin{algorithm}
	\KwData{this text}
	\KwResult{how to write algorithm with \LaTeX2e }
	initialization\;
	\While{not at end of this document}{
	read current\;
	\eIf{understand}{
	go to next section\;
	current section becomes this one\;
	}{
	go back to the beginning of current section\;
	}
	}
	\caption{How to write algorithms}
	\end{algorithm}
	\end{comment}
	\begin{figure*}
		\centering
		\small
		\scalebox{0.975}{
			\begin{tabular}{@{\hspace{0.0mm}}c@{\hspace{1.0mm}}c@{\hspace{1.0mm}}c@{\hspace{1.0mm}}c}
				\includegraphics[width=0.25\linewidth]{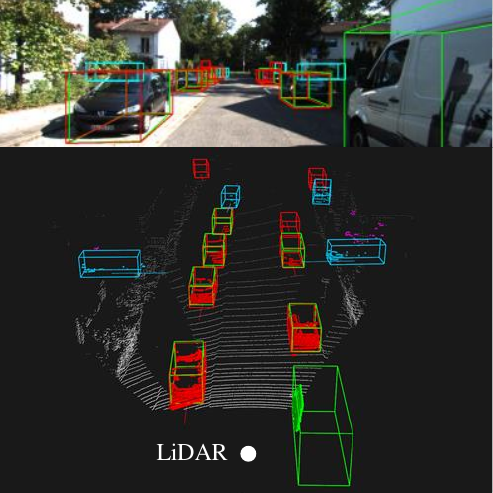}&
				\includegraphics[width=0.25\linewidth]{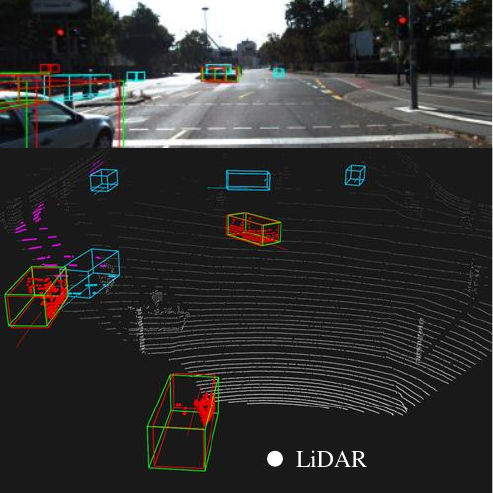}&
				\includegraphics[width=0.25\linewidth]{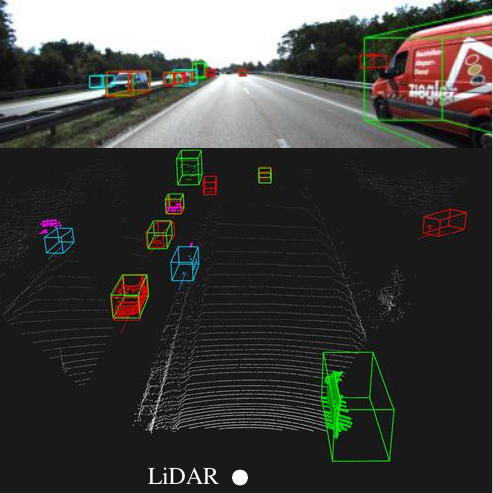}&
				\includegraphics[width=0.25\linewidth]{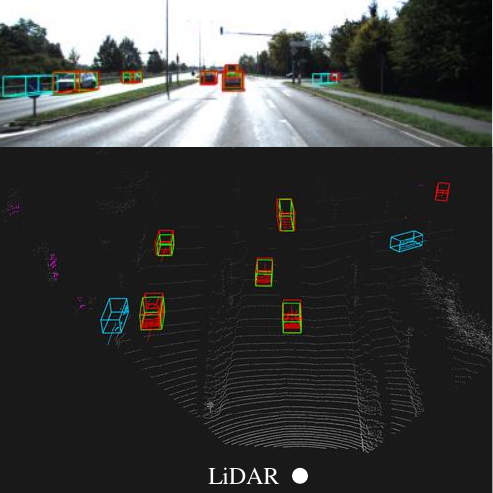}\\
		\end{tabular}}
		\caption{Qualitative results of PPC with PointRCNN on KITTI \emph{val} set. For each sample frame, 3D car detection result is visualized on the upper RGB image and in the lower point cloud. The ground-truth boxes are in green, the removed false positive boxes are in blue, the final predicted boxes are in red, and the penetrated points are in pink.
		}
		\label{fig:test_vis}
		\vspace{-0.3cm}
	\end{figure*}
	\vspace{0cm}
	\section{EXPERIMENTAL RESULT}
	\label{sec:exp}
	\vspace{-0.2cm}
	\subsection{Experiment Setup}
	\label{sec:expsu}
	\noindent
	{\bf Dataset}\hspace{0.3cm} We evaluate our module on challenging KITTI 3D car detection benchmark~\cite{geiger2012we}. The dataset contains 7,481 frames for training and 7,518 frames for testing, which includes images and point clouds. Since ground-truth labels are only provided in training set, we test our module on validation set. The training data is split into 3,769 frames and 3,712 frames for training and validation set as~\cite{chen2016monocular}.
	
	\smallskip
	\noindent
	{\bf Detection Network}\hspace{0.3cm} Among various detection networks, we exploit PointRCNN~\cite{shi2019pointrcnn} as the detector for our experiment, which is one of the state-of-the-art methods with high performance and publicly available as an open-source. 
	
	\smallskip
	\noindent
	{\bf CAD Point Cloud}\hspace{0.3cm} We select the CAD models of a car with generalized shapes, that are easily available on the Internet. The type of the cars used in our module is sedan, since its shape is included in that of most of the cars. In Table.~\ref{tab:model}, we prove that any car model of sedan without a unique feature can be used, by showing little difference in performances of PPC with different CAD models. 
	
	\begin{table}
		\small 
		\begin{center}
			\scalebox{1}[1]{
				\setlength\tabcolsep{10pt}
				\begin{tabular}{|c||cc|cc|}
					\hline
					\multirow{3}{*}{CAD Model} & 	
					\multicolumn{4}{c|}{~~Car - 3D Detection ~~}\\
					\cline{2-5}
					& \multicolumn{2}{c|}{HR-Precision} & \multicolumn{2}{c|}{Average Precision} \\
					%\cline{2-9}
					& Mod. & Hard & Mod. & Hard \\
					\hline
					Baseline & 46.59 & 53.90 & 82.26 & 77.99\\
					Model $1$ & 61.70 & 68.22 & 82.66 & 78.36\\
					Model $2$ & 63.34 & 69.66 & 82.58 & 78.39 \\
					%\rowcolor{LightCyan}
					Model $3$ & 62.36 & 68.81 & 82.54 & 78.37 \\
					Model $4$ & 62.05 & 68.53 & 82.65 & 78.36 \\
					\hline
			\end{tabular}	}
		\end{center}
		\vspace{-0.4cm}
		\caption{Performance comparison on the KITTI \textit{val} set using different CAD models with generalized shapes.
		}
		\label{tab:model}
	\end{table} 
	
	\begin{table}
		\small 
		\vspace{-2mm}
		\begin{center}
			\scalebox{0.85}[1]{
				\setlength\tabcolsep{10pt}
				\begin{tabular}{|c||cc|cc|}
					\hline
					\multirow{3}{*}{Method} & 	
					\multicolumn{4}{c|}{~~Car - 3D Detection ~~}\\
					\cline{2-5}
					& \multicolumn{2}{c|}{False Positive} & \multicolumn{2}{c|}{True Positive} \\
					%\cline{2-9}
					& Mod. & Hard & Mod. & Hard \\
					\hline
					Baseline & 18,610 & 18,594 & 123,870 & 153,709\\
					Baseline + PPC & 14,973 & 14,920 & 123,828 & 153,595\\
					\textit{Decrement} & \textit{-19.54\%} & \textit{-19.76\%} & \textit{-0.03\%} & \textit{-0.07\%} \\
					\hline
			\end{tabular}	}
		\end{center}
		\vspace{-0.2cm}
		\caption{Rate of changes in each sum of the number of false positive boxes and the number of true positive boxes on 40 recall positions.
		}
		\label{tab:fptp}
		\vspace{0cm}
	\end{table}

	\subsection{Evaluation and Discussion}
	\label{sec:eval}
	As we argue about the importance of the precision at high recall position in Sec.~\ref{sec:intro}, we mainly compute the precision at the highest recall rate possible for the detector, which will be termed \textit{HR-Precision}. Increment of the HR-Precision indicates the valid removal of false positive boxes without removing true positive boxes, while detecting as many ground-truth boxes as possible. The highest recall position of the PointRCNN is 87.5\% and 82.5\% in moderate and hard difficulty respectively. Average precision is also computed to demonstrate the overall performance of our module, as it is the evaluation metric generally used in detection. Note that the performance improvement in AP metric is not remarkable since our module focuses on sorting out the false positive boxes which are mostly generated in the quite high recall positions.
	
	We evaluate our method on both 3D car detection benchmark and Bird's Eye View (BEV) car detection benchmark for the moderate and hard difficulty, except the easy difficulty, as shown in Table.~\ref{tab:val}. For the case of the easy level, there is no significant improvement with our module since the number of false positive boxes is around 18\% compared to that of moderate and hard difficulty, which is too small to get a meaningful result. PPC increases the performance of the baseline detector, PointRCNN, with notable margins in HR-Precision for both moderate and hard level in 3D and BEV detection. Each sum of decrements of false positive boxes and true positive boxes on all recall positions is also computed in Table.~\ref{tab:fptp}. As seen in the table, about 20\% of false positive boxes are decreased, while about 99.9\% of true positive boxes are maintained. 
	
\begin{figure}[t]
		\begin{center}
			\includegraphics[width=1\linewidth]{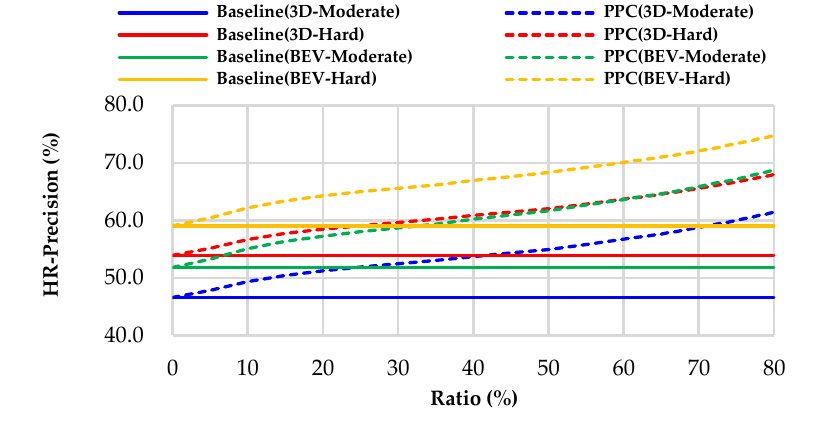}
			
		\end{center}
		\vspace{-0.6cm}
		\caption{Variation of 3D Detection HR-Precision (\%) on KITTI \textit{val} set. We plot the HR-Precision versus the ratio of the CAD model. The ratio of the CAD model indicates $\kappa$ in Eq.~\ref{eq1}.
		}
		\vspace{0cm}
		\label{fig:graph}
	\end{figure}

	We conduct the experiment for the ratio of CAD, which is described in Eq.~\ref{eq1} as $\kappa$. The reason $\kappa$ is essential is that the CAD model should not be fit exactly in the predicted box since the box with an overlap of over 70\% with the ground-truth box is regarded as the correct prediction. That is, the ratio $\kappa$ is an important parameter which prevents the correct prediction from being removed. Fig.~\ref{fig:graph} shows that the HR-Precision gets higher as the ratio of the CAD model increases. However, after 82\%, the highest recall position becomes lower since true positive boxes are removed due to the excessive ratio. Thus, 82\% is chosen for the ratio of the CAD model to prevent true positive boxes from being removed.
	
	\section{Conclusion}
	\vspace{-0.2cm}
	\label{sec:con}
	We have presented a novel approach for sorting out false positive boxes in 3D vehicle detection from LiDAR. The proposed method takes advantage of the obvious fact that the laser of LiDAR cannot penetrate the object. Our module exploits the CAD model of a car to replace the unobtainable shape of an actual object, and determines if any point is generated by penetrating the CAD model.
	Through the experiments, we show that this straightforward fact can have a vital role in classifying whether few points represent the object we search for, especially at high recall position.
	
	\smallskip
	\noindent
	\textbf{Acknowledgement.}\hspace{0.3cm}This work was supported by Institute for Information \& communications Technology Promotion (IITP) grant funded by the Korea government (MSIP). (No.2016-0-00197, Development of the high-precision natural 3D view generation technology using smart-car multi sensors and deep learning)
	
	
	\bibliographystyle{ieee}
	\bibliography{egbib}
\end{document}